\documentclass[a4paper,11pt]{article}
\usepackage{float}
\usepackage[pdftex]{graphicx}
\usepackage{subcaption}
\usepackage[T1]{fontenc}
\usepackage{lmodern}
\usepackage{slashed}
\usepackage[utf8]{inputenc}
\usepackage[english]{babel}
\usepackage{microtype}
\usepackage{cite}
\usepackage{amsmath,amssymb,amsfonts,amsthm}
\usepackage{mathtools,mathrsfs,calligra,aurical}
\usepackage[nottoc,notlot,notlof]{tocbibind}
\usepackage{upgreek}
\usepackage{mathtools}
\numberwithin{equation}{section}
\allowdisplaybreaks
\usepackage[all]{xy}
\usepackage{xcolor}
\usepackage{graphicx}
\graphicspath{{images/}}
\usepackage{geometry}
\usepackage[toc,page]{appendix}
\usepackage{hyperref}
\usepackage[normalem]{ulem}
\hypersetup{colorlinks = true, linkcolor = red, linktocpage = true, citecolor = blue}

\usepackage{bm}
\usepackage{ragged2e}
\usepackage{appendix}
\usepackage{slashed}
\usepackage{bbold}
\usepackage{cancel}

\newcommand{\be}{\begin{equation}}
\newcommand{\bea}{\begin{eqnarray}}

\newcommand{\ee}{\end{equation}}
\newcommand{\eea}{\end{eqnarray}}

\DeclareMathAlphabet{\mathpzc}{OT1}{pzc}{m}{it}

\usepackage{setspace}

\begin{document}

\begin{titlepage}
\begin{flushright}
\par\end{flushright}
\vskip 0.5cm
\begin{center}
\begin{spacing}{2.4}
\textbf{\huge  The Geometry of Meaning: Perfect Spacetime Representations of Hierarchical Structures}\\
\end{spacing}

\vskip 5mm

\vskip 1cm

\large {\bf Andr\'{e}s Anabal\'{o}n}\footnote{anabalo@gmail.com}
\vskip 1cm
\large {\bf Hugo Garc\'{e}s}\footnote{hugarces@inf.udec.cl} \large {\bf, Julio Oliva}\footnote{julioolivazapata@gmail.com}
\large {\bf, Jos\'{e} Cifuentes}\footnote{josecifuentesr@udec.cl}

\vskip .5cm 

{\textit{Departamento de F\'isica, Universidad de Concepci\'on, Casilla 160-C, Concepci\'on, Chile.}}
\vskip .5cm 
{\textit{Departamento Ingeniería Informática y Ciencias de la Computación, Universidad de Concepci\'on, Concepci\'on 4070409, Concepci\'on, Chile.}}
\vskip .5cm
{\textit{Departamento de Ingeniería Eléctrica, Universidad de Concepci\'on, Concepci\'on 4070386, Concepci\'on, Chile.}}
\end{center}

\vskip .5cm 
\begin{abstract}
We show that there is a fast algorithm that embeds hierarchical structures in three-dimensional Minkowski spacetime. The correlation of data ends up purely encoded in the causal structure. Our model relies solely on oriented token pairs---local hierarchical signals---with no access to global symbolic structure. We apply our method to the corpus of \textit{WordNet}. We provide a perfect embedding of the mammal sub-tree including ambiguities (more than one hierarchy per node) in such a way that the hierarchical structures get completely codified in the geometry and exactly reproduce the ground-truth. We extend this to a perfect embedding of the maximal unambiguous subset of the \textit{WordNet} with 82{,}115 noun tokens and a single hierarchy per token. We introduce a novel retrieval mechanism in which causality, not distance, governs hierarchical access. Our results seem to indicate that all discrete data has a perfect geometrical representation that is three-dimensional. The resulting embeddings are nearly conformally invariant, indicating deep connections with general relativity and field theory. These results suggest that concepts, categories, and their interrelations, namely hierarchical meaning itself, is geometric.

\end{abstract}

\vfill{}
\vspace{1.5cm}
\end{titlepage}

\setcounter{footnote}{0}
\section{Introduction}
Hierarchies pervade knowledge. In artificial intelligence—from knowledge graphs and ontologies to the structure of reasoning itself. They define relationships of generality, inheritance, and causality that are essential for abstraction and generalization. In symbolic systems, hierarchies enable efficient inference: knowing that a beagle is a type of dog and that dogs are mammals allows reasoning to transfer properties up and down the conceptual chain. However, in neural systems, these structures are often implicit, fragmented, or entirely absent. Embedding hierarchies into continuous geometric spaces offers a powerful unification: it allows symbolic structure to be modeled directly in a form that is differentiable, compositional, and compatible with gradient-based learning \cite{he2024language}.

In this paper, we introduce a new approach to representing hierarchical information based on causality—the intrinsic geometric structure of Minkowski spacetime. Causal ordering is the fundamental hierarchy governing physical reality; it underlies quantum field theory, general relativity, and string theory. As we show, it also provides the right framework for representing everyday human data.

The use cases are extensive. In language models, explicit hierarchical geometry could enable more robust generalization across levels of abstraction—e.g., understanding that “animal” subsumes “dog” and “cat” without requiring millions of co-occurrence examples. In knowledge representation, geometric embeddings offer scalable alternatives to symbolic graphs, where queries, analogies, and relational reasoning become operations in curved space \cite{park2024geometry}. In recommendation systems, product taxonomies and user behaviors follow implicit hierarchies; modeling these geometrically could lead to more coherent suggestions. In biomedicine, disease ontologies and protein function hierarchies already exist symbolically; embedding them in a structured geometric space could enable predictive models that respect domain logic. Even in scientific discovery, conceptual hierarchies—from particle physics to biology—form the backbone of theoretical structure, and geometric representations offer a path toward machine reasoning that mirrors scientific abstraction.

We report our results on a longstanding benchmark for hierarchical representation: the WordNet noun graph—a richly structured taxonomy that reaches depths of up to 20 nested “is-a” relations \cite{wordnet}. Although extensively studied, achieving a distortion-free geometric embedding of this hierarchy has remained an open challenge, with all existing approaches introducing approximation. Here, we introduce a method that yields a perfect embedding of a maximal unambiguous subset of WordNet, comprising 82,115 noun tokens and a single hierarchy per token, preserving every parent–child relation without distortion while remaining computationally efficient at scale. We extend our approach to achieve perfect accuracy even under structural ambiguities that typically confound hierarchical models. To test this, we provide a perfect embedding of the mammal subtree in the presence of ambiguities—namely, more than one hierarchy per token. Notably, the model is based solely on oriented token pairs—local relational data—without reliance on global symbolic priors.

Our method is extremely efficient in decreasing the number of dimensions of the representation. Modern AI, including large language models (LLMs), encodes knowledge in high-dimensional vectors (GPT-4 uses approximately 15,000 dimensions for its token embeddings). Yet the manifold hypothesis posits that meaningful structure in such data lies on a low-dimensional manifold, where relationships vary along only a few underlying directions. One promising route is to embed hierarchies in Euclidean geometries with non-trivial curvature. Hyperbolic embeddings place points in a negatively curved space that naturally reflects tree-like structure, capturing both similarity and hierarchical depth with far fewer dimensions than methods in flat space \cite{nickel2017poincare, nickel2018lorentz}. We emphasize that flat spacetimes—such as Minkowski space—can capture hierarchical relations not only through spatial intervals or distances, but also via causal (time) ordering. As we demonstrate below, leveraging both distance and causal structure enables the construction of perfect embeddings for highly complex data. Remarkably, this can be achieved not only with high efficiency but also in three-dimensional spaces. 

The structure of the paper is as follows. Since this work is intended for a broad audience, we begin by revisiting the simple and profound structure of nature that motivates our approach. We then describe the algorithm in detail and present our results. Finally, we conclude with a discussion of implications and directions for future research.

\section{A Natural Hierarchy}
Since Einstein's 1905 special relativity, our understanding of the structure of objective reality includes the idea of an intrinsic natural hierarchy, called causality. It is elegantly captured in its simplest instance by Minkowski spacetime, which provides the mathematical foundation for special relativity. The Minkowski metric that determines the invariant interval between points is given, in units where the speed of light is one, in its finite form by $\Delta s^2 = -\Delta t^2 + \Delta\vec{x} \cdot \Delta \vec{x}$. Here $\Delta$ represents simply the subtraction of the inertial coordinates of two points. Points in this spacetime are called events. This metric classifies relationships between events into three categories: timelike intervals ($\Delta s^2 < 0$), where events can be causally connected as signals traveling below light speed can connect them; lightlike intervals ($\Delta s^2 = 0$), where events are connected by light signals, defining the boundary of causal influence---the light cone; and spacelike intervals ($\Delta s^2 > 0$), where events cannot be causally connected, as they would require faster-than-light communication. This, together with the fact that all events connected by non-spacelike intervals are ordered in a unique way, invariant under the proper orthochronous Lorentz group, yields the causal structure of spacetime.

\begin{figure}[h!]
    \centering
    \includegraphics[width=0.7\textwidth]{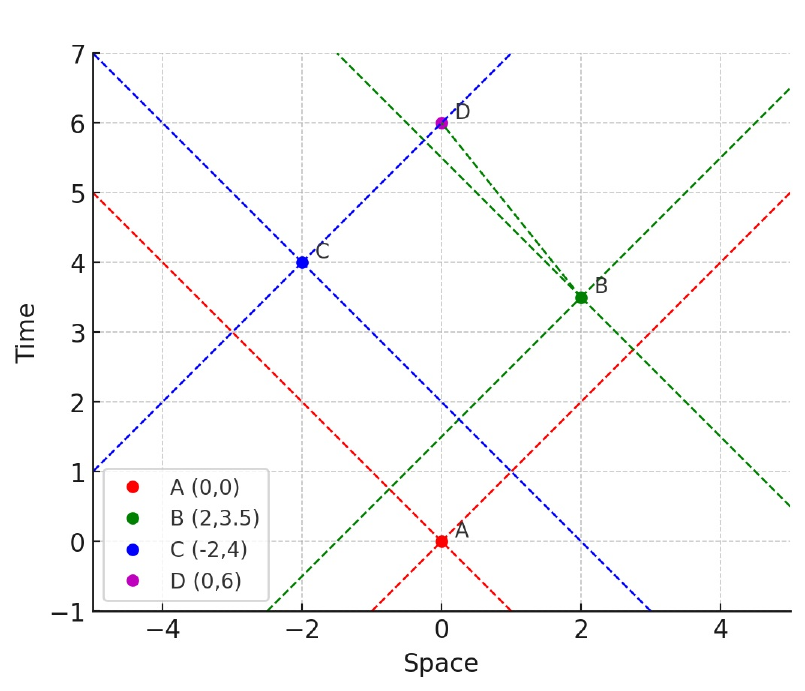}
    \caption{Events with their light cones. There are only two natural hierarchies in this figure $D \rightarrow C \rightarrow A$ and $B \rightarrow A$. Their transitive closure also follows from the causal diagram.}
    \label{fig:minkow}
\end{figure}

 In figure \ref{fig:minkow}, the four events are labeled $A$, $B$, $C$, and $D$. The interval between $C$ and $D$ is $0$. The interval between $B$ and $D$ is negative, which indicates that they influence each other. However, the first event that can exert influence on $D$ is $C$, and the first event that can influence $C$ is $A$. This is a very simple hierarchy that uses time ordering and interval size. Namely, consider all the points that are in the past light cone of an event. The one with the smallest $-\Delta s^2$ will be defined as the ancestor of that event. Now repeat the same process with the ancestor; that will yield the next member in the hierarchy and so on, until the last event, that one with no other event in its past light cone. We call this the natural hierarchy of events. Note that this process is local and only requires the information of the first neighborhood of each event, and therefore naturally includes the transitive closure of all hierarchies. Furthermore, it is bound to converge, as the root node will always have no event in its past.

 \section{The algorithm}
 To implement this idea, we begin by initializing the coordinates of events in each space dimension using a uniform distribution $\mathcal{U}(-1, 1)$. For the time dimension we initalize all the coordinates to $0$. Specifically, we associate one event to each token. We then consider sets of ordered pairs of the form $(X, Y)$, which can be derived, for example, from the transitive closure of a hierarchy, and have the interpretation that $X$ “is-a” $Y$. Let us denote the embeddings of these tokens by $(t_X, \vec{X})$ and $(t_Y, \vec{Y})$, respectively. We define the standard Euclidean distance between them as $D_{XY} = \sqrt{(\vec{X} - \vec{Y}) \cdot (\vec{X} - \vec{Y})}$, and their temporal order as $T_{XY} = t_X - t_Y$. After initialization, we identify the set of all violations to the causal relation as follows:
\begin{align}
\text{violations} = T_{XY} \leq 0\, \cup\,  T_{XY}  < D_{XY}
\end{align}
For each violation we define $T_{min} = D_{XY} + \epsilon_1$ and $\delta = T_{min} - T_{XY}$. Then, we adjust embeddings to enforce causality
\begin{align}
t_X \rightarrow t_X +\delta (1-\epsilon_2)\, , \qquad  t_Y \rightarrow t_Y - \delta \epsilon_2
\end{align}
This procedure is applied to the embedding of each pair of tokens in the dataset and iterated until convergence—that is, until the set of violations becomes empty. Importantly, only the time component of the embedding is updated; the spatial component remains fixed throughout. While generic values of the hyperparameters $\epsilon_1$ and $\epsilon_2$ typically lead to convergence, they do not yield high-quality embeddings.
In the experiments, we find good results by picking $\epsilon_1=10^{-5}$ and $\epsilon_2=0$. $\epsilon_2=0$ means that the coordinates of the $Y$ token are not updated. This implies that once the parent token is well adjusted respect to its children, only the children get further adjustments. The explanation for $\epsilon_1$ is more complex and we shall discuss it below.

After convergence, we compute the temporal ordering of all tokens and verify that, as expected, the root node corresponds to the smallest value of the time coordinate. For each token, we identify all other tokens lying within its past light cone and select as its ancestor the one connected by a geodesic with the smallest proper time—that is, the minimal value of $-\Delta s^2$. By applying this procedure iteratively to each token, we define a function that reconstructs the full causal hierarchy rooted at any token in the embedding. This hierarchical structure preserves the causal relationships encoded in the spacetime geometry, enabling us to trace chains of influence through the embedded representation.

It is important to emphasize that simply ordering the tokens by their proper time does not recover the correct hierarchy. In Figure \ref{fig:minkow}, the event $D$ is timelike connected to all other points, and the geodesics connecting $D$ to $(C, B, A)$ are ordered by increasing proper time. However, our method retrieves the hierarchy $D \rightarrow C \rightarrow A$, and $B$ does not appear in the hierarchy of $D$ at all. This highlights that proper time alone is insufficient to reconstruct the causal hierarchy inferred by our approach.

After convergence, the method typically yields perfect embeddings for approximately $90\%$ of the hierarchies—that is, $90\%$ of the tokens produce a hierarchy that exactly matches the ground truth\footnote{For batches of 10K tokens.}. The remaining $10\%$ lie within the causal structure but are not connected to their parent nodes via the geodesic of minimal proper time. To correct these, we iteratively revise the embedding using the token pairs and repair the misaligned child-parent pairs by connecting them by a randomly generated almost null geodesic. This geodesic is almost null because we require the proper time between parent and child to be of the order of $\epsilon_1$. This correction step relocates the misplaced child nodes and results in an algorithm that, in all our experiments, converges to a perfect embedding with exactly one hierarchy per token, fully reproducing the ground-truth.

Indeed, this approach works perfectly when there is only one hierarchical parent per token. However, in cases with multiple hierarchies, identifying the nearest token in the causal past of a child yields only a single parent. It is therefore natural to extend the notion of parenthood by considering not only the nearest but also the second-nearest token as potential parents. This procedure, however, would assign two parents to every token in the embedding, which is not always appropriate. To address this, we introduce a scale that quantifies proximity in the causal past, allowing us to discern when both candidates should be considered parents.  Specifically, by declaring that a token has two parents if and only if there exists a second token within a proximity of $10^{-6}=\frac{\epsilon_1}{10}$ from the primary parent, we achieve a precise separation between tokens with one parent and those with two. In the mammal sub-tree of WordNet, we found that this method performs exceptionally well and yields a perfect embedding in the presence of hierarchical ambiguity.

Note that this makes the value of $\epsilon_1$ relevant: it sets the scale at which nearest tokens can be unambiguously identified and it defines the introduction of a second, smaller scale to reliably detect the presence of multiple parents. 
 \section{Experiments}

\begin{figure}[h!]
    \centering
    \includegraphics[width=1.0 \textwidth]{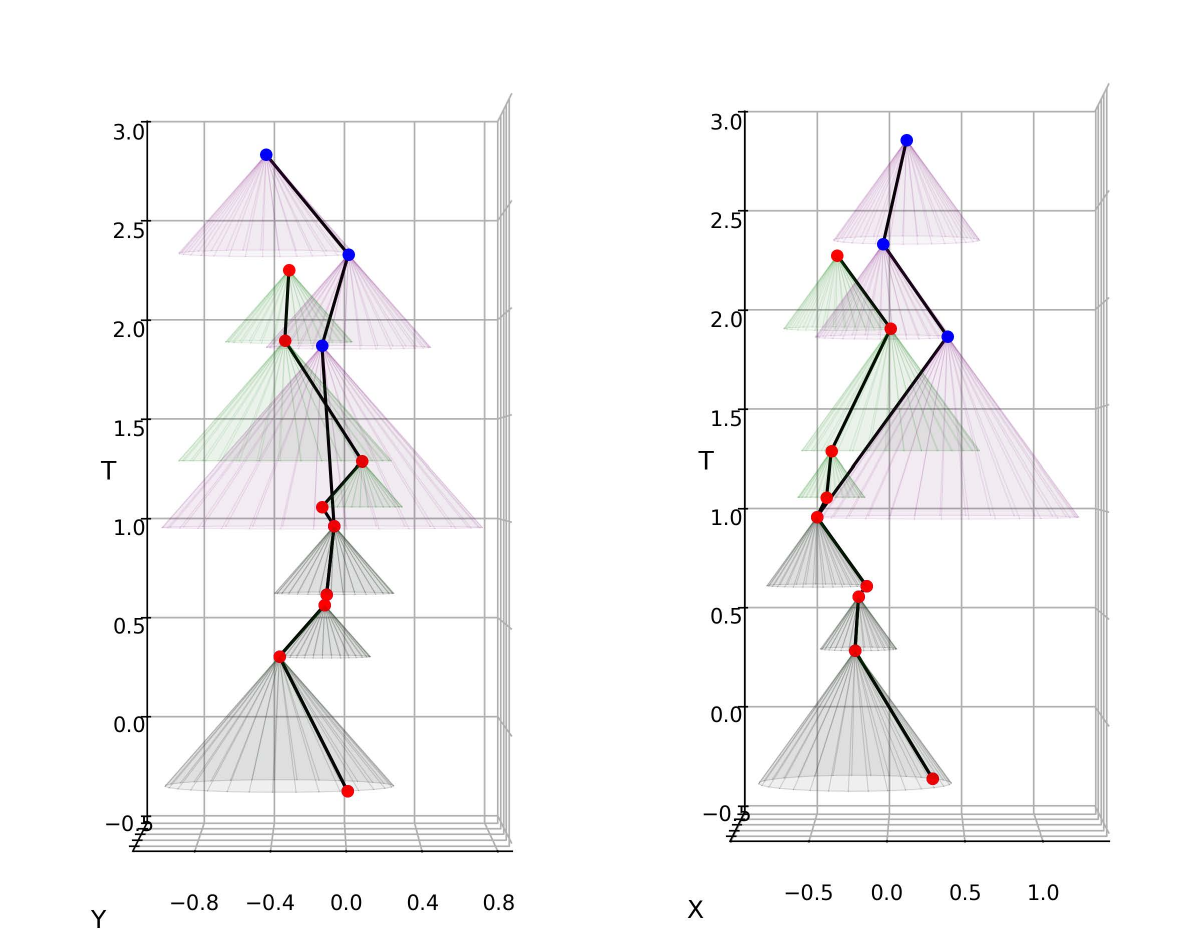}
    \caption{Left panel: Perfect embedding of two hierarchies — depth 8 (blue dots) and depth 9 (red dots) — both merging at the last 5 nodes. Past light cones are drawn for reference of the null surface at each token. It can be seen that geodesics connecting tokens are almost null. Right panel: Same embedding after a spatial rotation.}
    \label{fig:emb}
\end{figure}
WordNet contains \(82{,}115\) nouns with hierarchies reaching depths of up to \(20\). Among these, \(60{,}591\) nouns have unambiguous hierarchies. An additional \(17{,}553\) nouns possess exactly two hierarchies per leaf node. Cases with higher multiplicities become progressively rarer: \(1{,}281\) nouns have three, \(1{,}991\) have four, \(392\) have five, \(190\) have six, \(23\) have seven, \(69\) have eight, \(12\) have nine, another \(12\) have ten, and just one noun has twelve distinct hierarchies. Furthermore, the ambiguous hierarchies exhibit variability in depth. For example, the most ambiguous node can have depths in the closed interval $(8,13)$. In our experiments, we picked exactly one hierarchy per token. These hierarchies are structured such that each child has a unique parent across all hierarchies. Note that each parent may have multiple child tokens, as all hierarchies ultimately converge on the same token. The whole process of embedding all the correlations of this set perfectly in $D=3$ takes about 8 hours\footnote{The embedding itself takes about 30 minutes. Checking all the hierarchies and repairing them accounts for the remaining time, as fixing one hierarchy can disrupt another, and we verify all of them at each iteration.} in a computer with an Intel CPU Intel Core i7-7820X 3.6 GHz with $96$ GB of RAM without acceleration. The state-of-the-art on this dataset, using hyperbolic geometry, achieves a mean rank of $4.02$ and a MAP of $6.8$ in dimension $D = 10$ \cite{nickel2017poincare, nickel2018lorentz}. In contrast, our method achieves perfect retrieval, with both the mean rank and MAP equal to $1$. Since we curated WordNet to obtain these results, they are not directly comparable but serve only as a rough guide. The mammal sub-tree below offers a more appropriate setting for comparison.

This strongly suggests that the geometric representation of unambiguous hierarchies, namely where all the child tokens have a single parents across the system, has been solved in this paper.

The problem becomes significantly more challenging in the presence of ambiguities. In the WordNet noun graph, $2{,}213$ nodes have more than one parent---introducing all sources of hierarchical ambiguity. Of these, $1{,}737$ are leaf nodes, and $105$ are immediate parents of leaves. These nodes account for the entirety of the system’s structural ambiguity.

\begin{figure}[h!]
    \centering
    \includegraphics[width=0.9\textwidth]{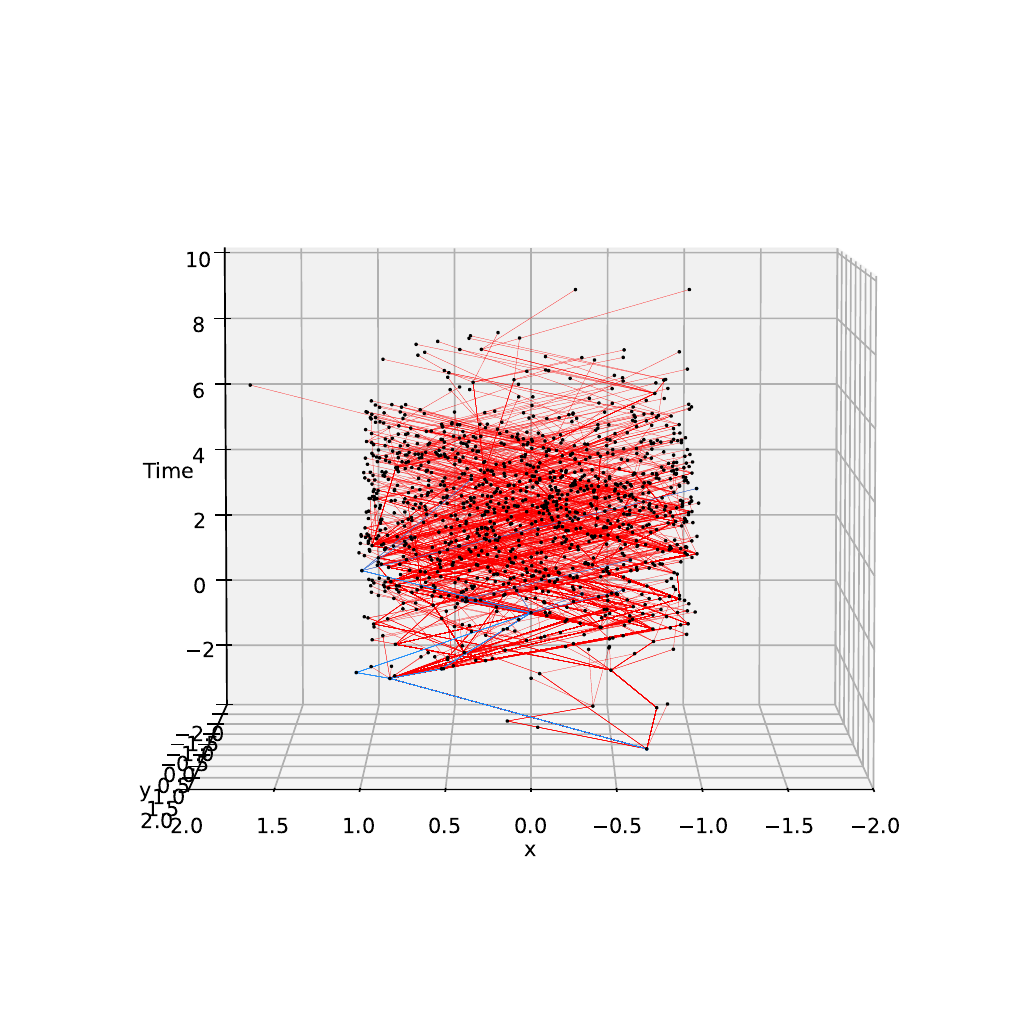}
    \caption{The figure shows a perfect embedding of the mammal sub‑tree. Whenever a token has a second hierarchy, that branch is highlighted in blue. The three‑dimensional layout offers a human‑friendly, visualization of the entire hierarchy.}
    \label{fig:emb2}
\end{figure}

To assess our approach in the presence of ambiguities but on a simpler subset, we begin with the smaller sub-tree rooted at ``mammal,'' which comprises $1{,}182$ tokens and $1{,}192$ hierarchies of depth up to $10$, including $20$ ambiguous cases. These ambiguities originate from $10$ nodes, each connected to the root by $2$ alternative hierarchical paths. The transitive closure of this structure yields $6{,}542$ node pairs, which we embed following the procedure described in the previous section. We fix the embedding dimension to $D=3$ and generate and generate the embedding over $1{,}182$ tokens with $1{,}192$ hierarchies. The whole ambiguity of this sub-tree is captured by a single child node that has two parents. We generate displacements of order $10^{-6}$ in the non-abiguous embedding until the time-like distance between the two parents is less than $10^{-6}$ and they are in the causal past of the child node. This lead to a geometric representation where our algorithm fully captures the existence of $1{,}192$ hierarchies identical to those of the ground-truth. The best results so far in the literature for this sub-tree are the Poincaré embedding with $D = 5$, which achieves a mean rank of $1.26$ and a MAP $0.927$ \cite{nickel2017poincare, nickel2018lorentz}. Ours is perfect. The perfect embedding takes approximately $1$ minute on a laptop with $16$ GB of RAM and a $12$th Gen Intel® Core™ i$9$-$12900$H processor running at $2.50$ GHz.
We draw some of the perfect embeddings in figure \ref{fig:emb}, where it can be seen that the representations have near null geodesics connecting the nodes within the hierarchy. The full mammal sub-tree can be found in \ref{fig:emb2}. They provide a novel visual representation of these hierarchies. 

 \section{Conclusions}
 
This breakthrough opens new possibilities for knowledge representation. We can equip LLMs with robust, interpretable concept spaces aligned to human ontologies and build semantic search engines that index information on a faithful hierarchical manifold for more precise retrieval. Even large biological taxonomies—such as the evolutionary tree of life—could be embedded geometrically without loss of detail, enabling analysis at unprecedented scale. Finally, this geometric approach suggests a solution to a known shortcoming of auto-regressive LLMs: compounding errors. Because these models generate text token by token, even a tiny per-token error rate makes the chance of an error-free sequence decline exponentially with sequence length. In effect, valid outputs form only a narrow branch in the vast tree of possible token sequences. By mapping this “valid subtree” as a manifold, we can constrain the model to navigate only along semantically valid paths. In practice, generation shifts from unguided token sampling to a guided traversal of the language network—preventing snowballing errors and keeping outputs factual \cite{dziri2023faith}.

This work proves that it is feasible to geometrically represent complex hierarchical data in three-dimensional Minkowski spacetime with perfect information retrieval, allowing for a new way of visualize complex data. Our experiments show that the representation is optimal when the geodesics within the representation are almost null (See figure \ref{fig:emb} ), resulting in an almost conformally invariant embedding. This suggests the possibility of generalization to representations of continuous data from the evolution of a conformal field in three dimensions, whose Green function has full support on the light cone.

Furthermore, our experiments yield remarkable results in three dimensions. This suggests that the dimensionality of any hierarchical discrete dataset is, at most, three—at least locally. Indeed, a knowledge tree with branching higher than two requires three dimensions to embed all branched nodes on the light cone—but no more than that. However, globally, more general geometries may be required. Our framework naturally extends beyond flat spacetime (i.e., Minkowski embeddings): the underlying causal structure can be generalized to include black holes, solitons, or wormholes. This opens a new bridge among general relativity, field theory, and information theory.

\section*{Acknowledgement}
AA is an Alexander von Humboldt Fellow. We are grateful for stimulating conversations with Mario Trigiante and Pietro Fré, and for Tomás Andrade’s comments on our manuscript. We also appreciate the encouragement to publish this work from Mariano Werner and Alice Martins‑Gistelinck. HG’s research is funded in part by FONDECYT Regular 1220903. This research was partially supported by the Patagón supercomputer at Universidad Austral de Chile (FONDEQUIP EQM180042) and by the Powered@NLHPC supercomputing infrastructure (CCSS210001). Some code snippets were drafted with the assistance of ChatGPT (OpenAI, model o3).Computational assistance provided by Claude 3.7 Sonnet (Anthropic, 2025). AA thanks Andrej Karpathy for his youtube channel.

\hypersetup{linkcolor=blue}
\phantomsection 
\addtocontents{toc}{\protect\addvspace{4.5pt}}
\bibliographystyle{mybibstyle}
\bibliography{biblio.bib}

\end{document}